\documentclass[letterpaper, 10 pt, conference, twoside]{ieeeconf}

\IEEEoverridecommandlockouts
\usepackage[T1]{fontenc}
\usepackage{times}
\usepackage{amsmath}
\usepackage{amssymb}
\usepackage{graphicx}
\usepackage[ruled,linesnumbered]{algorithm2e}
\usepackage{booktabs}
\usepackage{float}
\usepackage{bigstrut}
\usepackage{multirow}
\usepackage{balance}
\usepackage[hidelinks=true,bookmarks=false,colorlinks=true]{hyperref}
\usepackage{caption}
\usepackage{subcaption}
\usepackage{lineno}
\usepackage{amsopn}
\usepackage{comment}
\usepackage{color}
\usepackage{cite}
\usepackage{algorithmic}
\usepackage{tabularx}
\usepackage{xcolor}
\usepackage{mathrsfs}
\usepackage{makecell}

\makeatletter

\newcommand{\Rmnum}[1]{\expandafter\@slowromancap\romannumeral #1@}
\makeatother

\newcolumntype{L}[1]{>{\raggedright\arraybackslash}p{#1}}
\newcolumntype{C}[1]{>{\centering\arraybackslash}p{#1}}
\newcolumntype{R}[1]{>{\raggedleft\arraybackslash}p{#1}}

\usepackage{pifont}

\def\ie{\textit{i.e.}}
\def\eg{\textit{e.g.}}
\def\etal{\textit{et al.}}

\title{\LARGE \bf Learning Interpretable End-to-End Vision-Based Motion Planning for Autonomous Driving with Optical Flow Distillation}

\author{Hengli Wang, Peide Cai, Yuxiang Sun, Lujia Wang, and Ming Liu, \IEEEmembership{Senior Member, IEEE}
\thanks{\textit{(Corresponding author: Ming Liu.)}}
\thanks{Hengli Wang, Peide Cai and Ming Liu are with the Department of Electronic and Computer Engineering, The Hong Kong University of Science and Technology, Clear Water Bay, Kowloon, Hong Kong SAR, China (email: hwangdf@connect.ust.hk; pcaiaa@connect.ust.hk; eelium@ust.hk).}
\thanks{Yuxiang Sun is with the Department of Mechanical Engineering, The Hong Kong Polytechnic University, Hung Hom, Kowloon, Hong Kong (e-mail:
yx.sun@polyu.edu.hk, sun.yuxiang@outlook.com).}
\thanks{Lujia Wang is with Cloud Computing Lab of Shenzhen Institutes of Advanced Technology, Chinese Academy of Sciences, China (email: lj.wang1@siat.ac.cn).}
}

\begin{document}

\maketitle

\begin{abstract}
Recently, deep-learning based approaches have achieved impressive performance for autonomous driving. However, end-to-end vision-based methods typically have limited interpretability, making the behaviors of the deep networks difficult to explain. Hence, their potential applications could be limited in practice. To address this problem, we propose an interpretable end-to-end vision-based motion planning approach for autonomous driving, referred to as IVMP. Given a set of past surrounding-view images, our IVMP first predicts future egocentric semantic maps in bird's-eye-view space, which are then employed to plan trajectories for self-driving vehicles. The predicted future semantic maps not only provide useful interpretable information, but also allow our motion planning module to handle objects with low probability, thus improving the safety of autonomous driving. Moreover, we also develop an optical flow distillation paradigm, which can effectively enhance the network while still maintaining its real-time performance. Extensive experiments on the nuScenes dataset and closed-loop simulation show that our IVMP significantly outperforms the state-of-the-art approaches in imitating human drivers with a much higher success rate. Our project page is available at \url{https://sites.google.com/view/ivmp}.
\end{abstract}

\section{Introduction}
\label{sec.introduction}
Motion planning is an important capability in autonomous driving, serving as a fundamental building block \cite{liu2021hercules}. With the impressive advancement of deep learning technologies, many researchers have tried to develop end-to-end motion planning approaches using deep learning, which generally employ deep neural networks (DNNs) to directly map the raw sensor data (\eg, point clouds and images) to planned trajectories \cite{bansal2018chauffeurnet,cai2019vision,wang2021s2p2,cai2020vtgnet} or control commands (\eg, throttle and steering angle) \cite{bojarski2016end,gao2017intention,codevilla2018end}. However, end-to-end approaches are often criticized for their lack of interpretability. Here, interpretability refers to the ability to explain why the model can produce specific results \cite{sadat2020perceive}. Interpretability is very important for autonomous driving, since it can help people find out the limitations of the network and further improve it, especially when accidents such as collisions happen.

\begin{figure*}[t]
    \centering
    \includegraphics[width=0.9\textwidth]{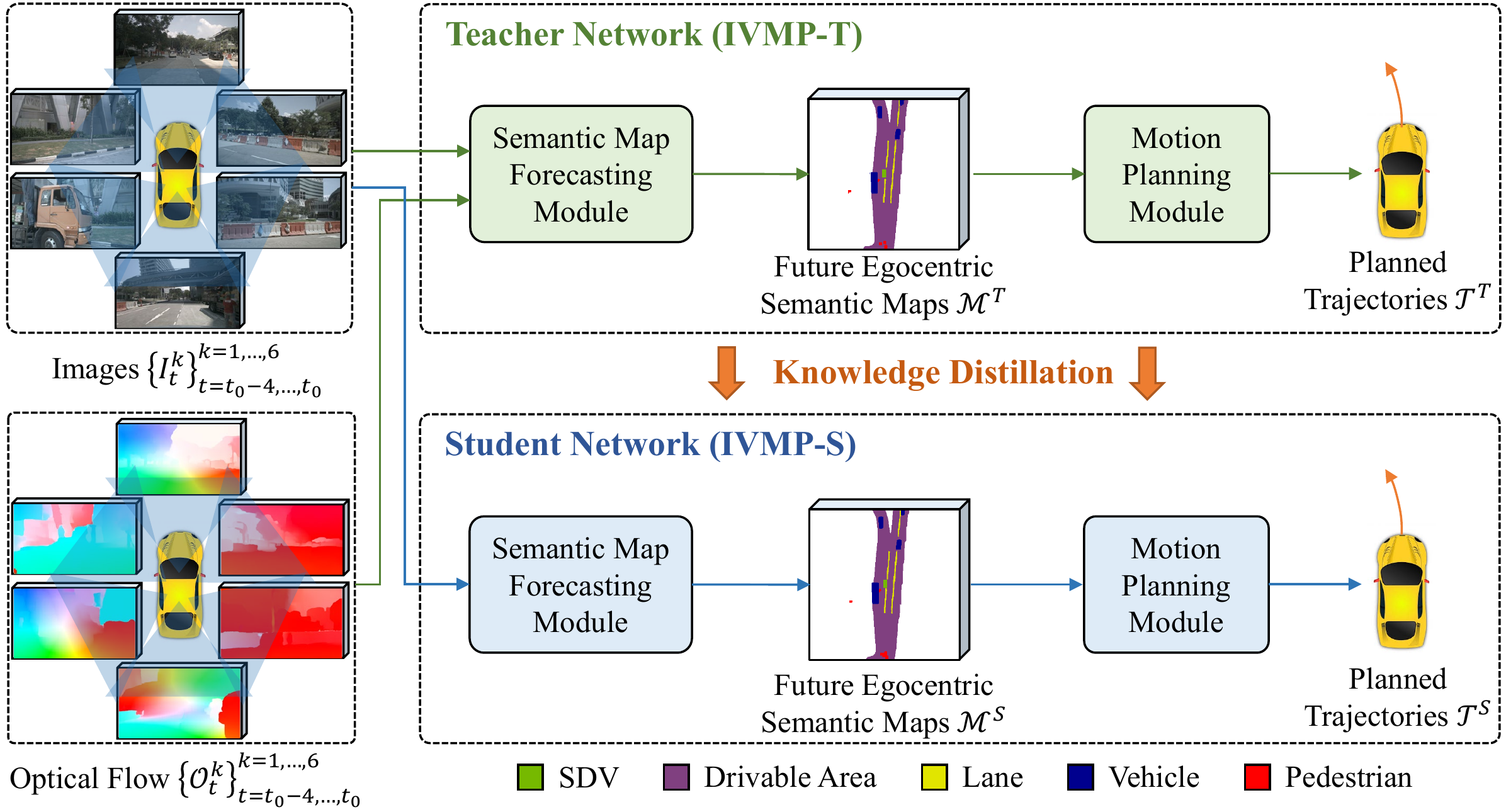}
    \caption{An overview of our IVMP, which consists of 1) a semantic map forecasting module to predict future egocentric semantic maps in BEV space and 2) a motion planning module to generate trajectories for SDVs. In the proposed optical flow distillation paradigm, the teacher network adopts a similar architecture to the student network but further takes optical flow information as input. We then use knowledge distillation techniques to effectively enhance the student network based on the teacher network, while still maintaining the real-time performance of the student network.}
    \label{fig.framework}
\end{figure*}

To improve the interpretability of end-to-end approaches, some researchers have adopted multi-task learning and developed models that can generate several interpretable representations, \eg, the object detection and prediction results \cite{zeng2020dsdnet} or the egocentric semantic maps in bird's-eye-view (BEV) space \cite{sadat2020perceive}. These approaches are generally based on LiDARs, since motion planning is usually performed in BEV space and the point clouds provided by LiDARs inherently meet this requirement. Unfortunately, images are located in perspective-view space, and there is a large gap between perspective-view space and BEV space. To mitigate this gap, some approaches have first transformed perspective images into BEV semantic maps, which are then employed to perform motion planning \cite{gupta2017cognitive}. However, the neglect of future environment prediction still restricts the interpretability of end-to-end vision-based approaches. Since images can provide more semantic information than point clouds, there is a strong motivation to improve the interpretability of end-to-end vision-based approaches.

In this paper, we propose an \underline{I}nterpretable end-to-end \underline{V}ision-based \underline{M}otion \underline{P}lanning approach, referred to as IVMP, for autonomous driving. Our IVMP, illustrated in Fig.~\ref{fig.framework}, takes as input a set of past surrounding-view images. We first lift these images into three dimensions (3-D) and employ a recurrent unit to predict a set of future egocentric semantic maps in BEV space. Afterwards, our motion planning module can employ these semantic maps to plan trajectories for the self-driving vehicle (SDV). Moreover, we develop an optical flow distillation paradigm to further improve the driving performance. Specifically, we refer to the above-mentioned network as the student network (IVMP-S) and additionally propose a teacher network (IVMP-T), which adopts a similar architecture to the student network but further takes optical flow information as input. The explicit motion information provided by the optical flow can significantly improve the teacher network, but the computation and corresponding feature processing of the optical flow also seriously slows down the network \cite{sun2018pwc,wang2020atg,wang2020cot}. We then use knowledge distillation techniques to effectively enhance the student network based on the teacher network, while still maintaining the real-time performance of the student network. To verify the effectiveness and efficiency of our approach, we perform evaluations on the popular nuScenes dataset \cite{caesar2020nuscenes}. In addition, we conduct closed-loop evaluations in the Carla simulation environment \cite{dosovitskiy2017carla}. The experimental results demonstrate that our approach can imitate human trajectories more closely than existing approaches with a much higher success rate. Furthermore, our student network runs much faster than the teacher network with similar driving performance thanks to the adopted optical flow distillation paradigm. The major contributions of this paper can be summarized as follows:
\begin{itemize}
    \item We propose IVMP, an interpretable end-to-end vision-based motion planner for autonomous driving.
    \item We develop an optical flow distillation paradigm, which can effectively enhance the network while still maintaining its real-time performance.
    \item We present extensive experiments on the nuScenes dataset and closed-loop simulation that demonstrate the effectiveness and efficiency of our IVMP.
\end{itemize}

\section{Related Work}
\label{sec.related_work}

\subsection{End-to-end Approaches for Autonomous Driving}
Traditional autonomous driving approaches generally perform motion planning based on the perception results \cite{wang2019self,wang2020applying,fan2020sne,wang2021pvstereo,fan2021learning,wang2021dynamic}, while end-to-end approaches directly map the raw sensor data to the planned trajectories or control commands. ALVINN was the first approach in this field, employing a 3-layer neural network to directly output control commands \cite{pomerleau1989alvinn}. Recently, with the success of deep learning, end-to-end approaches have advanced with deeper network architectures and more complex sensor inputs \cite{bojarski2016end,bansal2018chauffeurnet,wang2021s2p2,gao2017intention,codevilla2018end,cai2019vision,cai2020vtgnet}. However, these end-to-end approaches generally behave as black-box models and have limited interpretability as previously mentioned.

To improve the interpretability of end-to-end approaches, some researchers have adopted multi-task learning and developed LiDAR-based models that can generate several intermediate representations \cite{zeng2019end,zeng2020dsdnet,sadat2020perceive}. Specifically, Sadat \etal \cite{sadat2020perceive} employed LiDAR data to predict egocentric semantic maps in BEV space, which are then used for motion planning. Other researchers have followed this paradigm and attempted to improve the interpretability of end-to-end vision-based approaches. For example, Gupta \etal \cite{gupta2017cognitive} first used a multi-layer perceptron (MLP) to transform perspective images into BEV semantic maps, which are then employed to perform motion planning. However, their approach only utilizes a monocular camera with a limited field of view (FOV) and does not predict the future environment. In contrast, our approach takes a set of past surrounding-view images as input and predicts the future semantic maps in BEV space, which improves both interpretability and driving performance.

\begin{figure*}[t]
    \centering
    \includegraphics[width=0.99\textwidth]{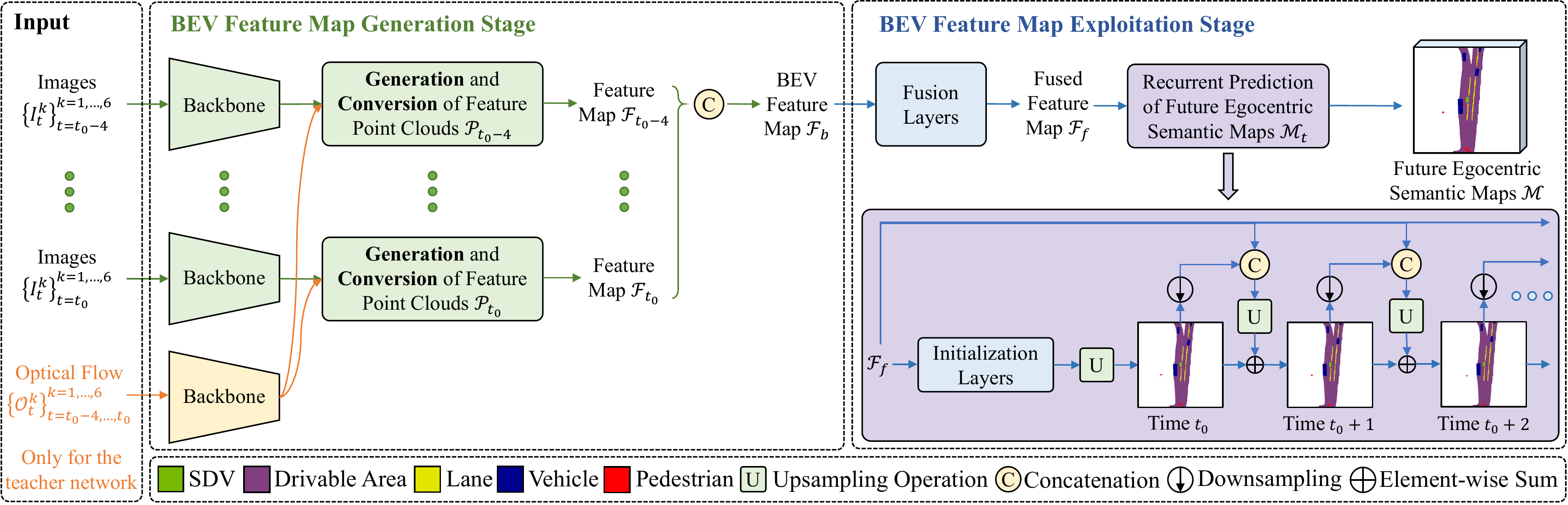}
    \caption{The pipeline of our semantic map forecasting module, which consists of a BEV feature map generation stage and a BEV feature map exploitation stage. In the first stage, we lift each image into 3-D to generate a feature map in BEV space, which is then processed by a recurrent unit to predict future egocentric semantic maps in the second stage. Please note that we omit the visualization of the softmax operation in the recurrent unit (purple color) for brevity.}
    \label{fig.module}
\end{figure*}

\subsection{Semantic Segmentation in BEV Space}
Many studies have used perspective images to perform semantic segmentation in BEV space \cite{reiher2020sim2real,lu2019monocular,pan2020cross,roddick2020predicting,philion2020lift}. Specifically, Pan \etal \cite{pan2020cross} employed an MLP to conduct such transformation for surrounding-view images. Roddick \etal \cite{roddick2020predicting} and Philion \etal \cite{philion2020lift} incorporated the strong geometric priors of camera extrinsic parameters into the pipeline, which presents impressive performance. Moreover, Deng \etal \cite{deng2019restricted} and Yang \etal \cite{yang2020omnisupervised} used fisheye images to generate semantic predictions in BEV space. The semantic map forecasting module in our approach is inspired by these works. The difference is that our approach has the capacity of predicting future semantic maps in BEV space.

\subsection{Knowledge Distillation}
Hinton \etal \cite{hinton2014distilling} first proposed the concept of knowledge distillation, which aims at leveraging the dark knowledge of a teacher network to train a student network with fewer parameters. Since then many techniques, such as Hint Training (HT) \cite{romero2015fitnets} and attention distillation \cite{zagoruyko2017paying}, have been developed to improve knowledge distillation, and it has been employed in many applications, \eg, semantic segmentation \cite{he2019knowledge} and object detection \cite{chen2017learning}. However, knowledge distillation for motion planning has not previously been explored, which is one of the major contributions of this paper.

\section{Methodology}
\label{sec.methodology}
In this section, we first introduce our semantic map forecasting module and motion planning module in Section~\ref{sec.semantic_map_prediction} and \ref{sec.motion_planning}, respectively. Then, we present our optical flow distillation paradigm in Section~\ref{sec.optical_flow_distillation_paradigm}. Finally, Section~\ref{sec.training_phase} elaborates the training phase.

\subsection{Semantic Map Forecasting Module}
\label{sec.semantic_map_prediction}
Let $I_{t}^{k} \in \mathbb{R}^{H \times W \times 3}$ denote the input RGB image, where $k = 1,\dots,6$ denotes the six cameras used in our experiments; and $t = t_0-4, \dots, t_0$ denotes the timestamp of the past five frames. The six cameras with known extrinsic parameters $\mathcal{E}^k$ and intrinsic parameters $\mathcal{I}^k$ roughly point in the forward, forward-left, forward-right, backward, backward-left and backward-right directions respectively. Then, given all images in the past five frames $\{I_{t}^{k}\}_{t=t_0-4, \dots, t_0}^{k=1,\dots,6}$, our semantic map forecasting module can output a set of egocentric semantic maps in the future eleven frames $\mathcal{M} = \{\mathcal{M}_t\}_{t=t_0,\dots,t_0+10}$. Fig.~\ref{fig.module} presents the pipeline of this module, which consists of a BEV feature map generation stage and a BEV feature map exploitation stage.

\subsubsection{BEV Feature Map Generation Stage}
The purpose of this stage is to lift each $I_{t}^{k}$ into 3-D to generate a feature map $\mathcal{F}_b$ in BEV space, which is the key to the prediction of $\mathcal{M}$. Inspired by \cite{philion2020lift}, we achieve this by generating contextual features at all possible depths for each pixel. Specifically, we associate each pixel with a set of $|D|$ discrete depths, where $D = \{d_0 + \Delta d, \dots, d_0 + |D|\Delta d\}$. Then, based on the intrinsic parameter $\mathcal{I}^k$, we can easily generate a large point cloud $\mathcal{P}_t^k$ that contains $H \cdot W \cdot |D|$ 3-D points for each $I_{t}^{k}$. The contextual feature for each point in $\mathcal{P}_t^k$ is a combination of the feature for the corresponding pixel and the discrete depth inference. To be specific, our network utilizes the backbone to predict a contextual feature $\mathbf{f} \in \mathbb{R}^{C}$ and a distribution $\pi$ over the discrete depth set $D$ for each pixel $\mathbf{p}$. The contextual feature $\mathbf{f}_d \in \mathbb{R}^{C}$ for point $\mathbf{p}_d$ is then computed by
\begin{equation}
    \mathbf{f}_d = \pi_d \cdot \mathbf{f},
    \label{eq.contextual_feature}
\end{equation}
where $d \in D$ refers to any discrete depth in $D$.

For the teacher network, we further incorporate optical flow into $\mathcal{P}_t^k$ to enable the network to better learn the past motion for predicting future semantic maps. Specifically, given any $I_{t}^{k}$ and $I_{t-1}^{k}$, we first utilize an off-the-shelf optical flow estimation network, PWCNet \cite{sun2018pwc}, to compute the backward optical flow $\mathcal{O}_{t}^{k} \in \mathbb{R}^{H \times W \times 2}$, which contains the explicit past motion information from $I_{t-1}^{k}$ to $I_{t}^{k}$. Then, we utilize another backbone to predict a contextual feature $\mathbf{f}' \in \mathbb{R}^{C}$ for each pixel $\mathbf{p}$. We concatenate $\mathbf{f}'$ with $\mathbf{f}$ and then generate a new feature. For notational simplicity, we still denote this new feature as $\mathbf{f}$. We further employ \eqref{eq.contextual_feature} to compute a contextual feature $\mathbf{f}_{d} \in \mathbb{R}^{C}$ for every point $\mathbf{p}_d$ in the teacher network. Note that the following architectures are almost the same for the teacher and student networks.

Then, for each timestamp $t$, we can utilize the extrinsic parameters $\{\mathcal{E}^{k}\}^{k=1,\dots,6}$ to aggregate $\{\mathcal{P}_{t}^{k}\}^{k=1,\dots,6}$ into a large point cloud $\mathcal{P}_{t}$. After that, we follow \cite{lang2019pointpillars} to convert $\mathcal{P}_{t}$ into ``pillars'', which refer to voxels with infinite height. Specifically, we assign each point to its nearest pillar and conduct pooling to construct a feature map $\mathcal{F}_{t} \in \mathbb{R}^{X' \times Y' \times C}$. Now $\mathcal{F}_{t}$ can be processed by convolutional layers to predict future egocentric semantic maps in BEV space. We then concatenate the features of all five past frames $\{\mathcal{F}_{t}\}_{t=t_0-4, \dots, t=t_0}$ to generate the BEV feature map $\mathcal{F}_{b}$.

\subsubsection{BEV Feature Map Exploitation Stage}
Given $\mathcal{F}_b$, which consists of all past information in BEV space, we will generate $\mathcal{M}$ in this stage. Note that $\mathcal{M}_{t} \in \mathbb{R}^{X \times Y \times |\mathcal{C}|}$, where $\mathcal{C}$ denotes the semantic classes, which include the drivable area, lane, vehicle and pedestrian in our experiments.

We first utilize several fusion layers to aggregate the spatio-temporal information of $\mathcal{F}_b$ and generate a fused feature map $\mathcal{F}_f$. The adopted fusion layers include two parallel convolutional layers with different dilation rates. Afterwards, we update the future egocentric semantic logits $\mathcal{S}_{t} \in \mathbb{R}^{X \times Y \times |\mathcal{C}|}$ repeatedly via a recurrent unit:
\begin{equation}
    \mathcal{S}_{t}(c) = \mathcal{S}_{t-1}(c) + \mathbf{U}\left(\mathbf{C}(\mathcal{F}_{f},\mathcal{S}_{t-1}(c)\downarrow)\right),
\end{equation}
where $c \in \mathcal{C}$ denotes any semantic class; $\downarrow$ denotes $\frac{1}{2} \times$ downsampling; $\mathbf{C}(\cdot,\cdot)$ denotes concatenation; and $\mathbf{U}(\cdot)$ denotes a $2\times$ upsampling operation, consisting of a $2\times$ bilinear interpolation followed by convolutional layers. Please note that $\mathcal{S}_{t_0}$ is predicted from $\mathcal{F}_{f}$ via initialization layers, which consist of convolutional layers and the upsampling operation $\mathbf{U}(\cdot)$. Then, we perform softmax on $\mathcal{S}_{t}$ to generate the future egocentric semantic map (predicted distribution) $\mathcal{M}_{t}$. We further define a future semantic map loss $\mathcal{L}_{M}$:
\begin{equation}
    \begin{aligned}
        \mathcal{L}_{M} (\widehat{\mathcal{M}}, \mathcal{M}) &= \sum_t H\left( \widehat{\mathcal{M}}_t, \mathcal{M}_t\right) \\
        &= - \sum_{t} \sum_{c}  \sum_{i,j} \widehat{\mathcal{M}}_t(i,j,c) \cdot \mathrm{log}( \mathcal{M}_t(i,j,c) ),
    \end{aligned}
    \label{eq.lm}
\end{equation}
where $H(\cdot,\cdot)$ denotes the cross entropy; and $\widehat{\mathcal{M}}$ denotes the ground truth distribution.

\subsection{Motion Planning Module}
\label{sec.motion_planning}
Based on $\mathcal{M}$, the current SDV state $\mathbf{s}_{t_0}$ and a given high-level route planned by a global planner, the purpose of our motion planning module is to generate a planned trajectory that contains the SDV states in the future ten frames, \ie, $\mathcal{T} = \{\mathbf{s}_t\}_{t=t_0+1,\dots,t_0+10}$. In our experiments, we adopt $\mathbf{s}_t = [x_t, y_t, \theta_t, \kappa_t, v_t, a_t]$, where $x$ and $y$ denote the position coordinates; and $\theta, \kappa, v$ and $a$ denote the heading angle, curvature, velocity and acceleration, respectively.

To achieve motion planning, we first employ the sampling technique proposed in \cite{sadat2019jointly} to sample a diverse set of trajectories for the SDV based on $\mathbf{s}_{t_0}$, and then select the one with the minimal cost of a learned cost function $f$ as follows:
\begin{equation}
    \mathcal{T}^{*} = \mathop{\arg\min}_{\mathcal{T}} f(\mathbf{s}_{t_0}, \mathcal{M}, \mathcal{T}; \mathbf{w}),
\end{equation}
where $\mathbf{w}$ denotes the learnable parameters of our motion planning module. $f$ consists of two subcosts: 1) $f_m$, which focuses on the safety of the planned trajectory based on $\mathcal{M}$; and 2) $f_o$, which focuses on the comfort and the consistency between the high-level route and the planned trajectory.

The intuition for $f_m$ is that the SDV should not collide with other objects and also should not drive on non-drivable areas. Thus, we define $f_m$ as follows:
\begin{equation}
    f_m = \sum_t \sum_c w_c \cdot \mathcal{M}_t(\mathcal{T}_t, c),
    \label{eq.fm}
\end{equation}
where $w_c \in \mathbf{w}$; $c \in \mathcal{C}'$ and $\mathcal{C}'$ includes the vehicle, pedestrian and non-drivable area (as opposed to the drivable area) classes in $\mathcal{M}$; and $\mathcal{M}_t(\mathcal{T}_t, c)$ denotes the corresponding probability for class $c$ on $\mathcal{M}_t$ based on the position provided by the sampled trajectory $\mathcal{T}_t$. The advantage of $f_m$ is that it employs the probability instead of the binary classification result, and thus can handle objects with low probability and further improve the safety of autonomous driving.

As for $f_o$, we define it as a linear combination of several cost terms. Specifically, to ensure the consistency between the high-level route and the planned trajectory, we adopt the distance between the end position of the trajectory and the given high-level route as a cost term. We also penalize the number of times the SDV changes lanes to encourage maneuvers that are consistent with the high-level route. Moreover, to encourage comfortable driving, we define several thresholds to penalize aggressive behaviors.

During training, considering that selecting the trajectory with the minimal cost in a discrete set is not differential, we follow \cite{sadat2020perceive} and develop a motion planning loss $\mathcal{L}_{P}$:
\begin{equation}
    \mathcal{L}_{P}(\widehat{\mathcal{T}}, \mathcal{T}) = \max_{\mathcal{T}} \left[ f(\widehat{\mathcal{T}}) - f(\mathcal{T}) + \sum_t ||\widehat{\mathcal{T}}_{t}-\mathcal{T}_{t}||_{1} \right]_+,
    \label{eq.lp}
\end{equation}
where $||\cdot||_1$ denotes the $L1$-Norm; $[\cdot]_+$ denotes the ReLU function; and $\widehat{\mathcal{T}}$ denotes the trajectory of human drivers. Please note that we omit $\mathbf{s}_{t_0}$, $\mathcal{M}$ and $\mathbf{w}$ in $f$ for brevity. $\mathcal{L}_{P}$ adopts a similar formulation to the max-margin loss, which can encourage the trajectories of human drivers to have a smaller cost $f$ than other trajectories. Moreover, $\mathcal{L}_{P}$ can also penalize trajectories that have a small cost but are different from the trajectories of human drivers.

\begin{figure*}[t]
    \centering
    \includegraphics[width=0.99\textwidth]{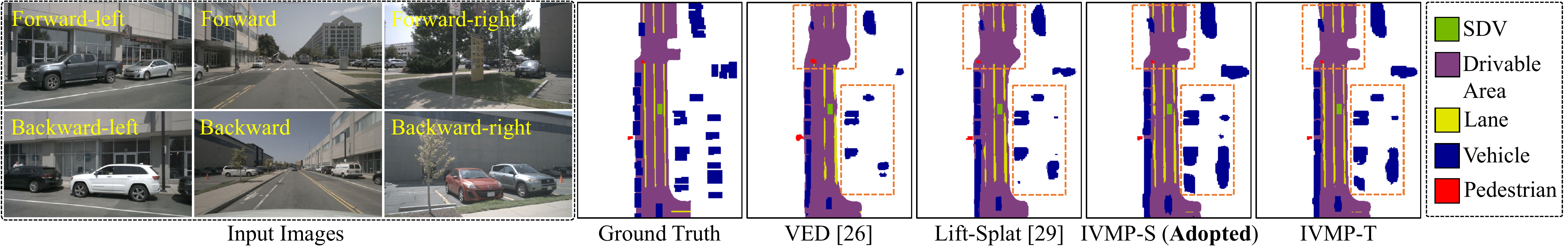}
    \caption{An example of the BEV semantic maps on the nuScenes dataset \cite{caesar2020nuscenes}. Please note that ``SDV'' does not belong to the semantic classes $\mathcal{C}$ and is only used for visualization. Significantly improved regions are marked with orange dashed boxes. }
    \label{fig.semantic}
\end{figure*}

\subsection{Optical Flow Distillation Paradigm}
\label{sec.optical_flow_distillation_paradigm}
Our teacher network and student network have been introduced in the above two subsections. The explicit motion information provided by the optical flow can significantly improve the teacher network, but the computation and corresponding feature processing of the optical flow also seriously slows down the network. In contrast, the student network can conduct motion planning in real time with a poorer performance than the teacher network. To further improve the driving performance of the student network, we develop an optical flow distillation paradigm, which distills the knowledge from a trained teacher network to the student network via knowledge distillation techniques. The distillation loss $\mathcal{L}_{D}$ consists of three terms, as follows:
\begin{equation}
    \mathcal{L}_{D} = \lambda_{DM} \mathcal{L}_{DM} + \lambda_{DP} \mathcal{L}_{DP} + \lambda_{DF} \mathcal{L}_{DF},
\end{equation}
where $\mathcal{L}_{DM}$, $\mathcal{L}_{DP}$ and $\mathcal{L}_{DF}$ denote the distillation loss for $\mathcal{M}$, $\mathcal{T}$ and $\mathcal{F}_b$, respectively; and $\lambda_{DM}$, $\lambda_{DP}$ and $\lambda_{DF}$ are hyperparameters that scale the three loss terms.

Specifically, since the prediction of future semantic maps is a classification task, $\mathcal{L}_{DM}$ is designed based on the conventional knowledge distillation technique \cite{hinton2014distilling}, as follows:
\begin{equation}
    \mathcal{L}_{DM} = \mathcal{L}_{M} (\mathcal{M}^{T}, \mathcal{M}^{S}) = \sum_t H\left( \mathcal{M}_t^T, \mathcal{M}_t^S \right),
\end{equation}
where $\mathcal{M}^{T}$ and $\mathcal{M}^{S}$ denote the predicted future semantic maps of the teacher and student network, respectively. Different from $\widehat{\mathcal{M}}$ in \eqref{eq.lm} that can only provide hard information, $\mathcal{M}^{T}$ can provide useful soft information. For example, for a pixel that belongs to the drivable area class, $\widehat{\mathcal{M}}$ can only show that it belongs to the drivable area class and does not belong to any other classes; while $\mathcal{M}^{T}$ can further show that it may belong to the lane class, but it is almost impossible for it to belong to the vehicle or the pedestrian classes. The soft information provided by $\mathcal{M}^{T}$ can effectively improve the student network.

In addition, inspired by \cite{chen2017learning}, we design $\mathcal{L}_{DP}$ as follows for motion planning:
\begin{equation}
    \nonumber
    \mathcal{L}_{DP}=\left\{
        \begin{array}{ll}
        \mathcal{L}_{P}(\mathcal{T}^{T*}, \mathcal{T}^{S}),\\ \hspace{0.4cm}\text {if} \sum_t ||\widehat{\mathcal{T}_{t}} - \mathcal{T}_{t}^{S*}||_{1} > \sum_t ||\widehat{\mathcal{T}_{t}} - \mathcal{T}_{t}^{T*}||_{1}, \\
        0, \hspace{0.1cm}\text {otherwise.}
    \end{array}\right.
\end{equation}
where $\mathcal{T}^{S}$ denotes a set of sampled trajectories of the student network; $\mathcal{T}^{T*}$ and $\mathcal{T}^{S*}$ denote the trajectories of the teacher and student network with the minimal cost of $f$, respectively; $\widehat{\mathcal{T}}$ denotes the trajectory of human drivers; and $\mathcal{L}_{P}(\cdot)$ is shown in \eqref{eq.lp}. $\mathcal{L}_{DP}$ encourages the student to be close to or better than the teacher, but does not push the student once it reaches the teacher's performance.

Since $\mathcal{F}_b$ of the teacher contains the explicit motion information provided by the optical flow while $\mathcal{F}_b$ of the student does not, we further develop $\mathcal{L}_{DF}$ based on HT \cite{romero2015fitnets}:
\begin{equation}
    \mathcal{L}_{DF} = ||\mathcal{F}_{b}^{T}-\mathcal{F}_{b}^{S}||_{1},
\end{equation}
where $\mathcal{F}_{b}^{T}$ and $\mathcal{F}_{b}^{S}$ denote the BEV feature map $\mathcal{F}_{b}$ of the teacher and student network, respectively. $\mathcal{L}_{DF}$ encourages the student network to mimic $\mathcal{F}_{b}$ of the teacher network.

\subsection{Training Phase}
\label{sec.training_phase}
In the training phase, we first utilize the following teacher training loss $\mathcal{L}^{T}$ to train the teacher network:
\begin{equation}
    \mathcal{L}^{T} = \lambda_{M} \mathcal{L}_{M}^{T} + \lambda_{P} \mathcal{L}_{P}^{T},
\end{equation}
where $\mathcal{L}_{M}^{T} = \mathcal{L}_{M} (\widehat{\mathcal{M}}, \mathcal{M}^T)$; $\mathcal{L}_{P}^{T} = \mathcal{L}_{P} (\widehat{\mathcal{T}}, \mathcal{T}^T)$; and $\lambda_{M}$ and $\lambda_{P}$ are hyperparameters that scale the loss terms.

Afterwards, we use the following student training loss $\mathcal{L}^{S}$ to train the student network based on the trained teacher network:
\begin{equation}
    \mathcal{L}^{S} = \lambda_{M} \mathcal{L}_{M}^{S} + \lambda_{P} \mathcal{L}_{P}^{S} + \lambda_{D} \mathcal{L}_{D},
\end{equation}
where $\mathcal{L}_{M}^{S} = \mathcal{L}_{M} (\widehat{\mathcal{M}}, \mathcal{M}^S)$; $\mathcal{L}_{P}^{T} = \mathcal{L}_{P} (\widehat{\mathcal{T}}, \mathcal{T}^S)$; and $\lambda_{M}$, $\lambda_{P}$ and $\lambda_{D}$ are hyperparameters that scale the loss terms.

\section{Experimental Results and Discussions}
\label{sec.experiment}

\subsection{Datasets and Implementation Details}
\label{sec.datasets_and_implementation_details}
In our experiments, we first use the nuScenes dataset \cite{caesar2020nuscenes} to evaluate the performance of our approach for BEV semantic map prediction and motion planning. We split the dataset into a training, a validation and a testing set that consist of 18072, 8019 and 8033 samples, respectively. Networks are first trained on the training set, then selected on the validation set and finally evaluated on the testing set. We also conduct closed-loop evaluation in the Carla simulation environment \cite{dosovitskiy2017carla}. Specifically, we first construct a large-scale driving dataset in different scenes, weather and illumination conditions. The dataset is then split into a training set with 200K samples and a validation set with 50K samples. Moreover, the closed-loop evaluation is performed in six scenes, including two unseen scenes. Each network is evaluated thoroughly with 1800 episodes (around 1000 km).

In the implementation, we adopt EfficientNet-B0 \cite{tan2019efficientnet} as the backbone. The time interval between two consecutive frames is $0.5s$, which means that our IVMP takes the information of the past $2s$ as input and generates planned trajectories for the future $5s$. We use the Adam optimizer \cite{kingma2014adam} with an initial learning rate of $10^{-4}$ to train our IVMP-T and IVMP-S on two NVIDIA GeForce RTX 2080 Ti GPUs. Moreover, we also train the student network without the proposed optical flow distillation paradigm, which is referred to as IVMP-S-ND, for better performance comparison.

\subsection{BEV Semantic Map Results on the nuScenes Dataset}
\label{sec.bev_semantic_map}

\begin{table}[t]
    \centering
    \caption{IoU (\%) results of BEV semantic maps on the nuScenes dataset \cite{caesar2020nuscenes}, where ``D'', ``L'', ``V'', ``P'' and ``M'' denote the drivable area, lane, vehicle, pedestrian and mean value, respectively. Best results are bolded.}
    \begin{tabular}{L{2.3cm}C{0.7cm}C{0.7cm}C{0.7cm}C{0.7cm}C{0.7cm}}
        \toprule
        Approach & D & L & V & P & M \\ \midrule
        VED \cite{lu2019monocular} & 60.82 & 16.74 & 23.28 & 11.93 & 28.19 \\
        VPN \cite{pan2020cross} & 65.97 & 17.05 & 28.17 & 10.26 & 30.36 \\
        PON \cite{roddick2020predicting} & 63.05 & 17.19 & 27.91 & 13.93 & 30.52 \\
        Lift-Splat \cite{philion2020lift} & 72.23 & 19.98 & 31.22 & 15.02 & 34.61 \\ \midrule
        IVMP-S-ND & 71.76 & 18.27 & 33.12 & 16.15 & 34.83 \\
        IVMP-S (\textbf{Adopted}) & 74.70 & 20.94 & 34.03 & \textbf{17.38} & 36.76 \\
        IVMP-T & \textbf{75.82} & \textbf{21.22} & \textbf{34.58} & 17.29 & \textbf{37.23} \\
        \bottomrule
    \end{tabular}
    \label{tab.eval_bev}
\end{table}

We adopt the intersection over union (IoU) as the evaluation metric, and the evaluation results of $\mathcal{M}_{t_0}$ are presented in Table~\ref{tab.eval_bev}. We can see that the three variants of our IVMP all outperform the state-of-the-art approaches, which demonstrates the effectiveness of our architecture that utilizes the past information. Moreover, IVMP-T achieves the best performance due to the explicit motion information provided by the optical flow. Furthermore, IVMP-S presents a much better performance than IVMP-S-ND and a similar performance to IVMP-T, which verifies the effectiveness of our optical flow distillation paradigm. The qualitative results in Fig.~\ref{fig.semantic} also confirm the above-mentioned conclusions. Please note that we adopt IVMP-S in practice due to its real-time performance. The analysis of inference time is presented in Section~\ref{sec.eval_motion_planning}.

\subsection{Motion Planning Results on the nuScenes Dataset}
\label{sec.eval_motion_planning}

\begin{figure}[t]
    \centering
    \includegraphics[width=0.8\linewidth]{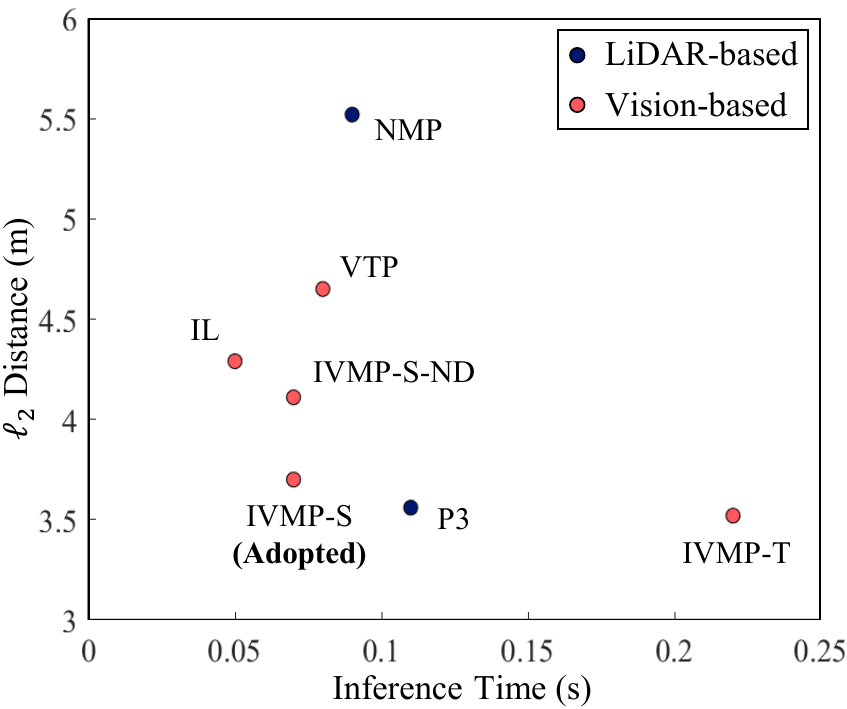}
    \caption{Motion planning results of IL, VTP \cite{cai2019vision}, P3 \cite{sadat2020perceive}, NMP~\cite{zeng2019end} and our IVMP on the nuScenes dataset \cite{caesar2020nuscenes}.}
    \label{fig.eval_mp}
\end{figure}

Following \cite{sadat2020perceive}, we use the $\ell_2$ distance between the planned trajectory and human trajectory at $t=5s$ for performance comparison. In addition, we also record the inference time of each approach. Fig.~\ref{fig.eval_mp} presents the evaluation results, where IL refers to an imitation learning baseline that predicts trajectories directly from $\mathcal{M}$. Please note that IL is also an end-to-end approach. From Fig.~\ref{fig.eval_mp}, we can clearly observe that the conclusions in Section~\ref{sec.bev_semantic_map} also hold for motion planning. Our IVMP-T achieves the most accurate performance due to the explicit motion information provided by the optical flow. Moreover, our IVMP-S can run in real time with a similar performance to IVMP-T thanks to the adopted optical flow distillation paradigm. In addition, one exciting fact is that our IVMP-S and IVMP-T can achieve competitive performance when compared to existing LiDAR-based approaches, which strongly demonstrates the effectiveness of our IVMP architecture with interpretable representations.

\subsection{Closed-loop Evaluation Results in the Carla Simulator}
\label{sec.eval_closed_loop}
\begin{figure}[t]
    \centering
    \includegraphics[width=0.99\linewidth]{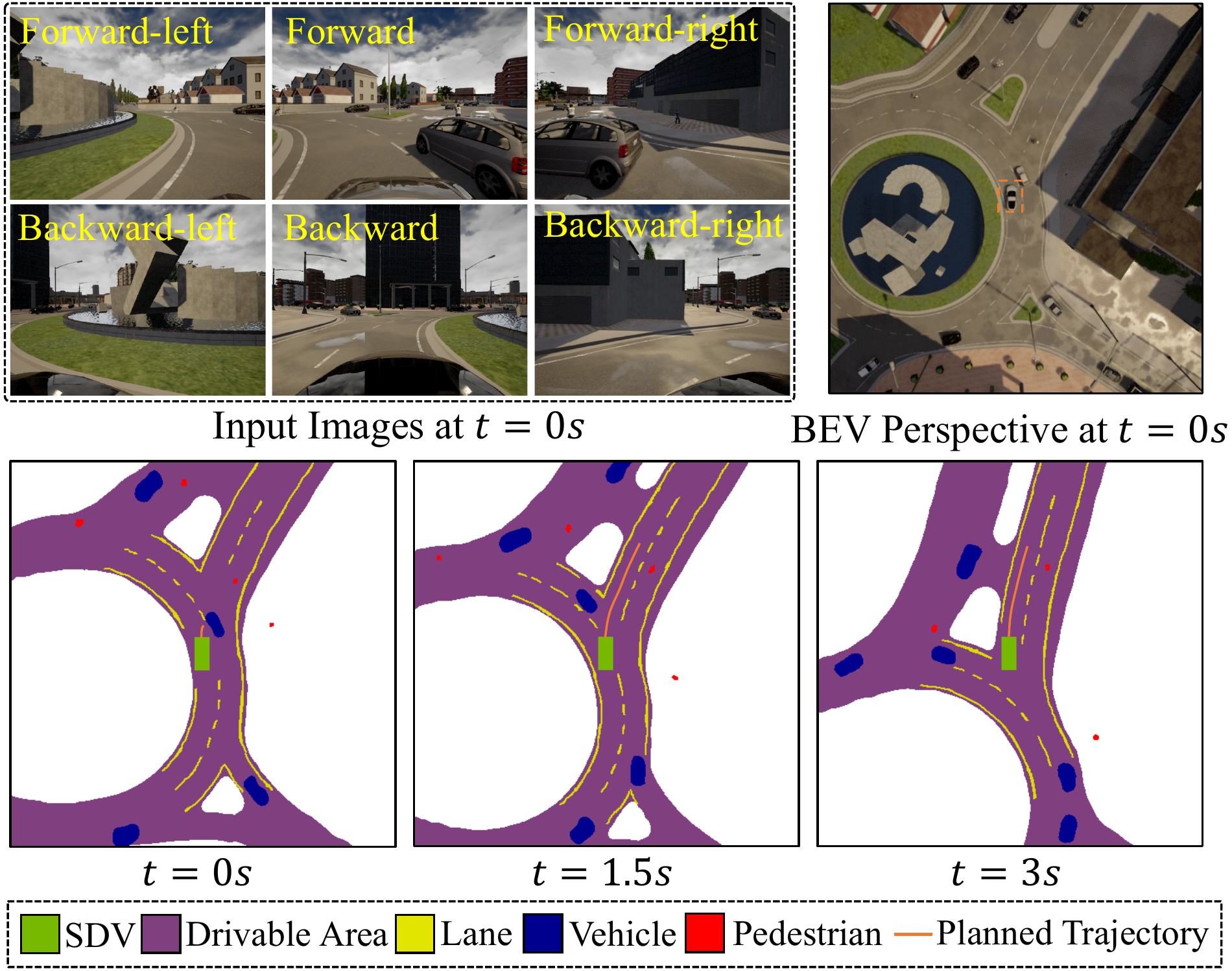}
    \caption{An example of the closed-loop evaluation in the Carla simulator \cite{dosovitskiy2017carla}. The SDV is marked with an orange dashed box in the BEV perspective.}
    \label{fig.planning}
    \vspace{-1em}
\end{figure}

Following \cite{cai2020probabilistic}, we adopt the success rate (SR) and right lane rate (RL) for evaluation. RL is defined as the proportion of the period in the given high-level route to the total driving time. We utilize a PID controller to transform the planned trajectories of our IVMP-S into control commands, and denote it as IVMP. We then compare the online performance of IVMP with Intention-Net \cite{gao2017intention} and CIL \cite{codevilla2018end}, as presented in Table~\ref{tab.carla}. We can observe that our IVMP achieves the best results in terms of both SR and RL. We analyze that the predicted semantic maps allow our motion planning module to handle objects with low probability, thus improving the safety of autonomous driving. Fig.~\ref{fig.planning} presents a driving scenario at intersections. Our IVMP can manuever the SDV to pass through the intersection safely and efficiently.

\begin{table}[t]
    \centering
    \caption{Closed-loop evaluation results in the Carla simulator \cite{dosovitskiy2017carla}. Best results are bolded.}
    \begin{tabular}{L{1.0cm}C{2.0cm}C{1.8cm}C{1.8cm}}
        \toprule
         & Intention-Net \cite{gao2017intention} & CIL \cite{codevilla2018end} & IVMP (\textbf{Ours}) \\ \midrule
        SR (\%) & 75.28 & 60.72 & \textbf{88.67} \\
        RL (\%) & 89.28 & 82.97 & \textbf{93.16} \\
        \bottomrule
    \end{tabular}
    \label{tab.carla}
\end{table}

\section{Conclusions}
\label{sec.conclusions}
In this paper, we proposed IVMP, an interpretable end-to-end vision-based motion planning approach for autonomous driving. Our IVMP first employs a semantic map forecasting module to predict future egocentric semantic maps in BEV space, which are then processed by a motion planning module to generate trajectories for SDVs. The predicted semantic maps not only provide useful interpretable information, but also allow our motion planning module to handle objects with low probability, thus improving the safety of autonomous driving. Moreover, we also develop an optical flow distillation paradigm, which can effectively enhance the network while still maintaining its real-time performance. Extensive experiments on the nuScenes dataset and closed-loop simulation have demonstrated the superiority of our IVMP over state-of-the-art approaches in BEV semantic map segmentation and imitating human drivers. We believe that our optical flow distillation paradigm can also be employed in other tasks related to spatio-temporal information analysis for performance improvement.

\clearpage
\bibliographystyle{IEEEtran}
\bibliography{egbib}

\begin{thebibliography}{10}
\providecommand{\url}[1]{#1}
\csname url@samestyle\endcsname
\providecommand{\newblock}{\relax}
\providecommand{\bibinfo}[2]{#2}
\providecommand{\BIBentrySTDinterwordspacing}{\spaceskip=0pt\relax}
\providecommand{\BIBentryALTinterwordstretchfactor}{4}
\providecommand{\BIBentryALTinterwordspacing}{\spaceskip=\fontdimen2\font plus
\BIBentryALTinterwordstretchfactor\fontdimen3\font minus
  \fontdimen4\font\relax}
\providecommand{\BIBforeignlanguage}[2]{{%
\expandafter\ifx\csname l@#1\endcsname\relax
\typeout{** WARNING: IEEEtran.bst: No hyphenation pattern has been}%
\typeout{** loaded for the language `#1'. Using the pattern for}%
\typeout{** the default language instead.}%
\else
\language=\csname l@#1\endcsname
\fi
#2}}
\providecommand{\BIBdecl}{\relax}
\BIBdecl

\bibitem{liu2021hercules}
T.~Liu, Q.~Liao \emph{et~al.}, ``The role of the {Hercules} autonomous vehicle
  during the {Covid-19} pandemic: An autonomous logistic vehicle for
  contactless goods transportation,'' \emph{IEEE Robotics and Automation
  Magazine}, 2021.

\bibitem{bansal2018chauffeurnet}
M.~Bansal, A.~Krizhevsky, and A.~Ogale, ``Chauffeurnet: Learning to drive by
  imitating the best and synthesizing the worst,'' \emph{arXiv preprint
  arXiv:1812.03079}, 2018.

\bibitem{cai2019vision}
P.~Cai, Y.~Sun, Y.~Chen, and M.~Liu, ``Vision-based trajectory planning via
  imitation learning for autonomous vehicles,'' in \emph{IEEE Intelligent
  Transportation Systems Conference (ITSC)}.\hskip 1em plus 0.5em minus
  0.4em\relax IEEE, 2019, pp. 2736--2742.

\bibitem{wang2021s2p2}
H.~Wang, Y.~Sun, R.~Fan, and M.~Liu, ``{S2P2}: Self-supervised goal-directed
  path planning using {RGB-D} data for robotic wheelchairs,'' in \emph{IEEE
  International Conference on Robotics and Automation}, 2021.

\bibitem{cai2020vtgnet}
P.~Cai, Y.~Sun, H.~Wang, and M.~Liu, ``{VTGNet}: A vision-based trajectory
  generation network for autonomous vehicles in urban environments,''
  \emph{IEEE Transactions on Intelligent Vehicles}, 2020.

\bibitem{bojarski2016end}
M.~Bojarski, D.~Del~Testa, D.~Dworakowski, B.~Firner, B.~Flepp, P.~Goyal, L.~D.
  Jackel, M.~Monfort, U.~Muller, J.~Zhang \emph{et~al.}, ``End to end learning
  for self-driving cars,'' \emph{arXiv preprint arXiv:1604.07316}, 2016.

\bibitem{gao2017intention}
W.~Gao, D.~Hsu, W.~S. Lee, S.~Shen, and K.~Subramanian, ``Intention-net:
  Integrating planning and deep learning for goal-directed autonomous
  navigation,'' in \emph{Conference on Robot Learning (CoRL)}, 2017.

\bibitem{codevilla2018end}
F.~Codevilla, M.~Miiller, A.~L{\'o}pez, V.~Koltun, and A.~Dosovitskiy,
  ``End-to-end driving via conditional imitation learning,'' in \emph{2018 IEEE
  International Conference on Robotics and Automation (ICRA)}.\hskip 1em plus
  0.5em minus 0.4em\relax IEEE, 2018, pp. 1--9.

\bibitem{sadat2020perceive}
A.~Sadat, S.~Casas, M.~Ren, X.~Wu, P.~Dhawan, and R.~Urtasun, ``Perceive,
  predict, and plan: Safe motion planning through interpretable semantic
  representations,'' in \emph{Proceedings of the European Conference on
  Computer Vision (ECCV)}, 2020.

\bibitem{zeng2020dsdnet}
W.~Zeng, S.~Wang, R.~Liao, Y.~Chen, B.~Yang, and R.~Urtasun, ``{DSDNet}: Deep
  structured self-driving network,'' in \emph{Proceedings of the European
  Conference on Computer Vision (ECCV)}, 2020.

\bibitem{gupta2017cognitive}
S.~Gupta, J.~Davidson, S.~Levine, R.~Sukthankar, and J.~Malik, ``Cognitive
  mapping and planning for visual navigation,'' in \emph{Proceedings of the
  IEEE Conference on Computer Vision and Pattern Recognition (CVPR)}, 2017, pp.
  2616--2625.

\bibitem{sun2018pwc}
D.~Sun, X.~Yang, M.-Y. Liu, and J.~Kautz, ``{PWC-Net: CNNs} for optical flow
  using pyramid, warping, and cost volume,'' in \emph{Proceedings of the IEEE
  Conference on Computer Vision and Pattern Recognition (CVPR)}, 2018, pp.
  8934--8943.

\bibitem{wang2020atg}
H.~Wang, Y.~Liu, H.~Huang, Y.~Pan, W.~Yu, J.~Jiang, D.~Lyu, M.~J. Bocus,
  M.~Liu, I.~Pitas \emph{et~al.}, ``{ATG-PVD}: Ticketing parking violations on
  a drone,'' in \emph{European Conference on Computer Vision}.\hskip 1em plus
  0.5em minus 0.4em\relax Springer, 2020, pp. 541--557.

\bibitem{wang2020cot}
H.~Wang, R.~Fan, and M.~Liu, ``{CoT-AMFlow}: Adaptive modulation network with
  co-teaching strategy for unsupervised optical flow estimation,'' \emph{arXiv
  preprint arXiv:2011.02156}, 2020.

\bibitem{caesar2020nuscenes}
H.~Caesar, V.~Bankiti, A.~H. Lang, S.~Vora, V.~E. Liong, Q.~Xu, A.~Krishnan,
  Y.~Pan, G.~Baldan, and O.~Beijbom, ``{nuScenes}: A multimodal dataset for
  autonomous driving,'' in \emph{Proceedings of the IEEE/CVF Conference on
  Computer Vision and Pattern Recognition (CVPR)}, 2020, pp. 11\,621--11\,631.

\bibitem{dosovitskiy2017carla}
A.~Dosovitskiy, G.~Ros, F.~Codevilla, A.~Lopez, and V.~Koltun, ``Carla: An open
  urban driving simulator,'' in \emph{Conference on Robot Learning (CoRL)},
  2017.

\bibitem{wang2019self}
H.~Wang, Y.~Sun, and M.~Liu, ``Self-supervised drivable area and road anomaly
  segmentation using rgb-d data for robotic wheelchairs,'' \emph{IEEE Robotics
  and Automation Letters}, vol.~4, no.~4, pp. 4386--4393, 2019.

\bibitem{wang2020applying}
H.~Wang, R.~Fan, Y.~Sun, and M.~Liu, ``Applying surface normal information in
  drivable area and road anomaly detection for ground mobile robots,'' in
  \emph{2020 IEEE/RSJ International Conference on Intelligent Robots and
  Systems (IROS)}, 2020.

\bibitem{fan2020sne}
R.~Fan, H.~Wang, P.~Cai, and M.~Liu, ``{SNE-RoadSeg}: Incorporating surface
  normal information into semantic segmentation for accurate freespace
  detection,'' in \emph{European Conference on Computer Vision}.\hskip 1em plus
  0.5em minus 0.4em\relax Springer, 2020, pp. 340--356.

\bibitem{wang2021pvstereo}
H.~Wang, R.~Fan, P.~Cai, and M.~Liu, ``{PVStereo}: Pyramid voting module for
  end-to-end self-supervised stereo matching,'' \emph{IEEE Robotics and
  Automation Letters}, 2021.

\bibitem{fan2021learning}
R.~Fan, H.~Wang, P.~Cai, J.~Wu, M.~J. Bocus, L.~Qiao, and M.~Liu, ``Learning
  collision-free space detection from stereo images: Homography matrix brings
  better data augmentation,'' \emph{IEEE Transactions on Mechatronics}, 2021.

\bibitem{wang2021dynamic}
H.~Wang, R.~Fan, Y.~Sun, and M.~Liu, ``Dynamic fusion module evolves drivable
  area and road anomaly detection: A benchmark and algorithms,'' \emph{IEEE
  Transactions on Cybernetics}, 2021.

\bibitem{pomerleau1989alvinn}
D.~A. Pomerleau, ``Alvinn: An autonomous land vehicle in a neural network,'' in
  \emph{Advances in Neural Information Processing Systems (NIPS)}, 1989, pp.
  305--313.

\bibitem{zeng2019end}
W.~Zeng, W.~Luo, S.~Suo, A.~Sadat, B.~Yang, S.~Casas, and R.~Urtasun,
  ``End-to-end interpretable neural motion planner,'' in \emph{Proceedings of
  the IEEE Conference on Computer Vision and Pattern Recognition (CVPR)}, 2019,
  pp. 8660--8669.

\bibitem{reiher2020sim2real}
L.~Reiher, B.~Lampe, and L.~Eckstein, ``A sim2real deep learning approach for
  the transformation of images from multiple vehicle-mounted cameras to a
  semantically segmented image in bird's eye view,'' in \emph{2020 IEEE
  Intelligent Transportation Systems Conference (ITSC)}, 2020.

\bibitem{lu2019monocular}
C.~Lu, M.~J.~G. van~de Molengraft, and G.~Dubbelman, ``Monocular semantic
  occupancy grid mapping with convolutional variational encoder--decoder
  networks,'' \emph{IEEE Robotics and Automation Letters}, vol.~4, no.~2, pp.
  445--452, 2019.

\bibitem{pan2020cross}
B.~Pan, J.~Sun, H.~Y.~T. Leung, A.~Andonian, and B.~Zhou, ``Cross-view semantic
  segmentation for sensing surroundings,'' \emph{IEEE Robotics and Automation
  Letters}, vol.~5, no.~3, pp. 4867--4873, 2020.

\bibitem{roddick2020predicting}
T.~Roddick and R.~Cipolla, ``Predicting semantic map representations from
  images using pyramid occupancy networks,'' in \emph{Proceedings of the
  IEEE/CVF Conference on Computer Vision and Pattern Recognition (CVPR)}, 2020,
  pp. 11\,138--11\,147.

\bibitem{philion2020lift}
J.~Philion and S.~Fidler, ``Lift, splat, shoot: Encoding images from arbitrary
  camera rigs by implicitly unprojecting to 3d,'' in \emph{Proceedings of the
  European Conference on Computer Vision (ECCV)}, 2020.

\bibitem{deng2019restricted}
L.~Deng, M.~Yang, H.~Li, T.~Li, B.~Hu, and C.~Wang, ``Restricted deformable
  convolution-based road scene semantic segmentation using surround view
  cameras,'' \emph{IEEE Transactions on Intelligent Transportation Systems},
  vol.~21, no.~10, pp. 4350--4362, 2019.

\bibitem{yang2020omnisupervised}
K.~Yang, X.~Hu, Y.~Fang, K.~Wang, and R.~Stiefelhagen, ``Omnisupervised
  omnidirectional semantic segmentation,'' \emph{IEEE Transactions on
  Intelligent Transportation Systems}, 2020.

\bibitem{hinton2014distilling}
G.~Hinton, O.~Vinyals, and J.~Dean, ``Distilling the knowledge in a neural
  network,'' in \emph{Advances in Neural Information Processing Systems
  Workshop (NIPSW)}, 2014.

\bibitem{romero2015fitnets}
A.~Romero, N.~Ballas, S.~E. Kahou, A.~Chassang, C.~Gatta, and Y.~Bengio,
  ``Fitnets: Hints for thin deep nets,'' in \emph{International Conference on
  Learning Representations (ICLR)}, 2015.

\bibitem{zagoruyko2017paying}
S.~Zagoruyko and N.~Komodakis, ``Paying more attention to attention: Improving
  the performance of convolutional neural networks via attention transfer,'' in
  \emph{International Conference on Learning Representations (ICLR)}, 2017.

\bibitem{he2019knowledge}
T.~He, C.~Shen, Z.~Tian, D.~Gong, C.~Sun, and Y.~Yan, ``Knowledge adaptation
  for efficient semantic segmentation,'' in \emph{Proceedings of the IEEE
  Conference on Computer Vision and Pattern Recognition (CVPR)}, 2019, pp.
  578--587.

\bibitem{chen2017learning}
G.~Chen, W.~Choi, X.~Yu, T.~Han, and M.~Chandraker, ``Learning efficient object
  detection models with knowledge distillation,'' in \emph{Advances in Neural
  Information Processing Systems (NIPS)}, 2017, pp. 742--751.

\bibitem{lang2019pointpillars}
A.~H. Lang, S.~Vora, H.~Caesar, L.~Zhou, J.~Yang, and O.~Beijbom,
  ``Pointpillars: Fast encoders for object detection from point clouds,'' in
  \emph{Proceedings of the IEEE Conference on Computer Vision and Pattern
  Recognition (CVPR)}, 2019, pp. 12\,697--12\,705.

\bibitem{sadat2019jointly}
A.~Sadat, M.~Ren, A.~Pokrovsky, Y.-C. Lin, E.~Yumer, and R.~Urtasun, ``Jointly
  learnable behavior and trajectory planning for self-driving vehicles,'' in
  \emph{Proceedings of the IEEE/RSJ International Conference on Intelligent
  Robots and Systems (IROS)}, 2019.

\bibitem{tan2019efficientnet}
M.~Tan and Q.~V. Le, ``Efficientnet: Rethinking model scaling for convolutional
  neural networks,'' in \emph{International Conference on Machine Learning
  (ICML)}, 2019.

\bibitem{kingma2014adam}
D.~P. Kingma and J.~Ba, ``Adam: A method for stochastic optimization,''
  \emph{arXiv preprint arXiv:1412.6980}, 2014.

\bibitem{cai2020probabilistic}
P.~Cai, S.~Wang, Y.~Sun, and M.~Liu, ``Probabilistic end-to-end vehicle
  navigation in complex dynamic environments with multimodal sensor fusion,''
  \emph{IEEE Robotics and Automation Letters}, vol.~5, no.~3, pp. 4218--4224,
  2020.

\end{thebibliography}

\end{document}